\documentclass[cameraready]{vgtc}                 

\graphicspath{{figures/}{pictures/}{images/}{./}} 

\usepackage{times}                     

\usepackage{tabu}                      
\usepackage{booktabs}                  
\usepackage{lipsum}                    
\usepackage{mwe}                       
\usepackage[normalem]{ulem}
\usepackage{mathptmx}                  
\usepackage{amssymb} 
\usepackage{booktabs} 
\usepackage{multirow}
\usepackage{array}
\usepackage{bm}
\usepackage[ruled]{algorithm2e} 

\SetAlFnt{\small}
\SetAlCapFnt{\small}
\SetAlCapNameFnt{\small}
\SetAlCapHSkip{0pt}

\newcolumntype{C}[1]{>{\centering\arraybackslash}m{#1}}

\usepackage{xcolor}
\definecolor{aliceblue}{rgb}{0.26, 0.48, 0.89}
\definecolor{bobgreen}{rgb}{0.0, 0.5, 0.0}
\definecolor{charliered}{rgb}{0.85, 0.1, 0.1}
\definecolor{pink}{rgb}{1, 0, 1}

\vgtccategory{Research}

\vgtcinsertpkg

\title{Parameter-Free \textcolor{black}{Neural Lens} Blur Rendering for High-Fidelity Composites}

\usepackage{orcidlink} 
\author{
Lingyan Ruan\,\orcidlink{0000-0001-7799-9148}\thanks{e-mail: lingyan.ruan@unimelb.edu.au} %
\and Bin Chen\,\orcidlink{0000-0003-3022-1931}\thanks{e-mail: bin.chen@unimelb.edu.au} %
\and Taehyun Rhee\,\orcidlink{0000-0002-6150-0637}\thanks{e-mail: taehyun.rhee@unimelb.edu.au (corresponding author)}
}
\affiliation{School of Computing and Informatoin Systems, University of Melbourne}

\teaser{
  \centering
  \includegraphics[width=\linewidth]{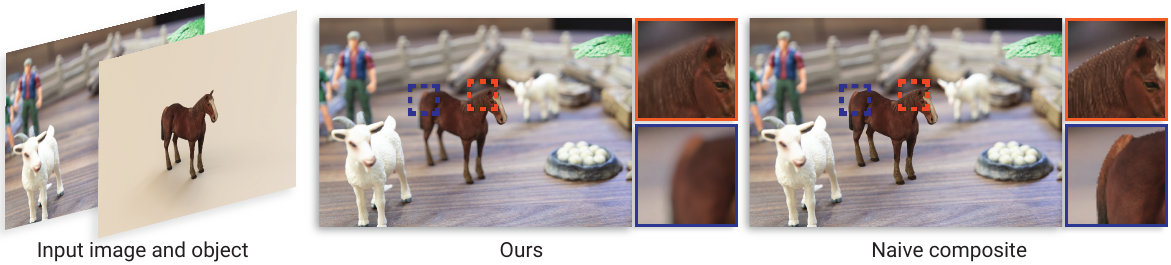}
  \caption{Our approach enables natural and consistent augmentation of images with virtual objects, preserving realistic Depth-of-Field (DoF) effects without requiring prior information such as camera metadata or scene depth. It supports spatially varying blur, as demonstrated by the virtual horse: its head remains in focus while the tail appears out of focus, consistent with the background.}
  \label{fig:teaser}
}

\abstract{

Consistent and natural camera lens blur is important for seamlessly blending 3D virtual objects into photographed real-scenes. Since lens blur typically varies with scene depth, the placement of virtual objects and their corresponding blur levels significantly affect the visual fidelity of mixed reality compositions. Existing pipelines often rely on camera parameters (e.g., focal length, focus distance, aperture size) and scene depth to compute the circle of confusion (CoC) for realistic lens blur rendering. However, such information is often unavailable to ordinary users, limiting the accessibility and generalizability of these methods. In this work, we propose a novel compositing approach that directly estimates the CoC map from RGB images, bypassing the need for scene depth or camera metadata. The CoC values for virtual objects are inferred through a linear relationship between its signed CoC map and depth, and realistic lens blur is rendered using a neural reblurring network. Our method provides flexible and practical solution for real-world applications. Experimental results demonstrate that our method achieves high-fidelity compositing with realistic defocus effects, outperforming state-of-the-art techniques in both qualitative and quantitative evaluations.

}


\begin{document}

\firstsection{Introduction}

\maketitle

Realistic compositing of 3D virtual objects into real-world photographs is a fundamental objective in Mixed Reality (MR), as it directly influences users' visual perception and immersive experience. While extensive research has been dedicated to achieving photorealistic rendering by modeling scene geometry, surface reflectance, and illumination conditions \cite{fournier1993common,alhakamy2020real,chalmers2014perceptually,aittala2010inverse,pilet2006all}, the visual appearance of captured photographs is also significantly affected by various camera-induced artifacts, such as motion blur, defocus blur, sensor noise, vignetting, and lens distortion \cite{klein2009simulating}. To ensure visual coherence in the composited photographs, many existing methods apply a degradation process to the rendered virtual objects, mimicking the camera-induced effects so that the virtual content can blend naturally into the scene \cite{okumura2006augmented,prakash2025blind,mandl2021neural,mandl2024neural}. Among these effects, defocus blur, which is often associated with a shallow depth of field, is of particular interest due to its prominence in consumer photography and its intentional use for aesthetic purposes. Because it varies spatially with scene depth, it presents additional challenges for blending 3D virtual objects seamlessly into real environments. Although several techniques have been proposed to simulate lens blur, they typically require special calibration \cite{klein2009simulating} or the presence of markers within the scene \cite{okumura2006augmented,kan2012physically}. Such requirements are rarely met in practical scenarios, making these methods less suitable for general real-world applications where camera parameters are unknown or inaccessible.

Recent advances such as Neural Camera \cite{mandl2021neural} and Neural Bokeh \cite{mandl2024neural} address these limitations by introducing neural network-based solutions for achieving visual coherence. The Neural Camera approach uses multilayer perceptrons to model complex camera effects, including depth of field, color response, and sensor noise, by fitting to specific camera characteristics. Neural Bokeh focuses on learning realistic bokeh effects to enhance the integration of virtual content into real scenes. However, both methods often rely on access to specific camera metadata or calibration data, which limits their applicability in situations where such information is unavailable, such as in images or videos collected from the internet. To overcome these challenges, our work proposes generalizable method that removes the dependency on camera parameters while maintaining high quality in visual compositing. The most closely related work is the recent approach by Prakash et al. \cite{prakash2025blind}, which accounts for motion blur, depth of field, and sensor noise in video compositing for mixed reality. Although their method also does not require camera metadata, it models depth of field using quadratic function of scene depth, which may not accurately represent the characteristics of real optical blur. Our method shares the same motivation of enabling high quality compositing without camera specific data, but aims to improve both the realism and the general applicability of defocus blur modeling.

\begin{itemize}
\vspace{-0.1cm}
\item We present Neural Lens, the first method to directly estimate the CoC map for image compositing, bypassing the need for explicit disparity estimation, camera parameters, or hardware calibration data. This enables a more flexible, robust, and generalizable compositing framework.
\vspace{-0.1cm}
\item We specifically address the inference of CoC values for virtual objects by leveraging the linear relationship between the signed CoC of the photograph scene and the disparity of virtual objects, thereby resolving the inherent scale ambiguity between real and virtual content.
\vspace{-0.1cm}
\item We comprehensively evaluate our approach on both rendered and real-world photographed scenes. In addition, we conduct user studies to assess the perceived naturalness and visual realism of the composited results.

\end{itemize}

\section{Related Work}

\subsection{Depth of field Synthesis}
\textcolor{black}{Depth of Field (DoF) refers to the range of scene depths that appear sharp in an image, while regions outside this range appear increasingly blurred. A common approach to simulating DoF involves computing a per-pixel Circle of Confusion (CoC) map based on the thin lens model, which quantifies the blur radius as a function of depth or disparity (scaled reciprocal of depth). This process typically begins by estimating the scene depth or disparity, and then combining it with camera metadata such as aperture size and focal length to compute the CoC for each pixel. The resulting spatially varying CoC map is then used to guide the lens blur rendering across the image.} In this section, we review CoC estimation and lens reblurring techniques.

\noindent \textbf{CoC Map Estimation:} The CoC value at a given pixel quantifies the size of the blur circle for the scene point imaged at that pixel \cite{potmesil1981lens}. As shown in Fig. \ref{fig:coc}, the CoC size is determined by the point’s depth relative to the focal plane and the camera’s internal parameters (focal length, aperture size, sensor size, etc.). For synthesizing a DoF effect on an all-in-focus image, one typically begins by estimating a monocular depth or disparity map \cite{ranftl2020towards,Vitor2023towards}. Given this estimated depth and the known camera settings (focal length, aperture, sensor size) and chosen focus distance, a CoC map can be derived by converting each depth value to an appropriate blur radius. CoC map is also often represented as defocus map that encodes the per-pixel blur amount; for example, existing works \cite{lee2019deep,gur2019single,ruan2021aifnet} represent the defocus level by Gaussian kernel standard deviation, set to a value proportional to the CoC. Recent learning-based approaches have improved CoC and defocus estimation by jointly modeling depth and blur. Zhang and Sun \cite{zhang2021joint} and Piché-Meunier et al.\cite{piche2023lens} propose networks that simultaneously estimate the depth map and defocus blur from a single defocused image while enforcing physical consistency constraints between them. More recently, Ruan et al. \cite{ruan2024self} further advanced CoC estimation by directly predicting the CoC map from a single image via a transfer learning strategy. Their approach bypasses the need for an explicit intermediate depth map.

This work focuses on a more generalized and robust method for estimating the CoC map from a single image. In the following Section \ref{sec:analysis_sota}, we discuss the prevailing trends in CoC estimation, including methods that derive CoC from depth maps as well as approaches that predict the CoC map directly, to position our approach in the context of the state of the art.

\noindent \textbf{Lens Blur Rendering:} Lens blur effects typically require an RGB-D input and can be simulated using either object-space or image-space techniques. Object-space methods physically simulate the camera's optics by tracing rays through a lens model, producing highly realistic lens blur effects \cite{lee2010real,yu2010real}, but it is computationally expensive and requires a complete 3D scene, which limits its practicality for real-world images. Image-space methods, in contrast, apply defocus as a post-processing effect on a 2D image using depth information. Traditional image-space techniques (i.e. classical methods) blur each pixel according to its depth via depth-dependent convolution or splatting operations \cite{zhang2019synthetic, wadhwa2018synthetic}, which are far more efficient than ray tracing, but they often suffer from artifacts at depth discontinuities where foreground and background pixels blend incorrectly. Recently, neural rendering approaches have been explored to overcome some limitations of classical techniques. Instead of manually designing a blur filter, these methods train deep neural networks to directly generate a shallow depth-of-field image from an input photo and its depth/CoC map \cite{xiao2018deepfocus,wang2018deeplens,mandl2024neural}. However, a drawback of purely learned approaches is the lack of explicit control and physical fidelity. \textcolor{black}{Hybrid methods combining classical and neural techniques have been proposed to leverage the strengths of both. BokehMe \cite{peng2022bokehme} and their expanded work BokehMe++ \cite{peng2024bokehme++} fuses a physics-based renderer with a neural network to produce photo-realistic bokeh effects. In contrast, Dr.Bokeh \cite{sheng2024dr} introduces a fully differentiable, occlusion-aware rendering pipeline. In this work, we adopt BokehMe \cite{peng2022bokehme} for its simplicity and strong performance in our setting.}

\subsection{Coherent Lens Blur Rendering for Compositing}
Visual consistency between synthetic objects and real photographs significantly affects the quality of image composites and user perception \cite{collins2017visual}. This consistency is especially critical in augmented virtual rendering within mixed reality (MR) systems. As discussed in \cite{fournier1993common,alhakamy2020real,chalmers2014perceptually,aittala2010inverse,pilet2006all}, achieving such consistency typically requires estimating factors like illumination, geometry, viewing pose, and reflectance. In our work, we assume these factors are known and focus on the lens blur effect.

Camera lens effects in augmented reality has been studied over the past decades. This includes phenomena such as defocus blur \cite{okumura2006augmented,prakash2025blind,mandl2021neural,mandl2024neural}, motion blur \cite{klein2009simulating,fischer2006enhanced,park2009esm,okumura2006augmented,prakash2025blind}, color-space conversion \cite{klein2009simulating}, chromatic aberration \cite{klein2009simulating} and camera noise \cite{fischer2006enhanced,prakash2025blind,mandl2021neural}. In this work, we specifically focus on lens blur consistency in image and video composites. Early work \cite{okumura2006augmented} addressed depth-of-field (DoF) and motion blur by fitting a point spread function (PSF) to real images based on a circular AR marker in the scene. Synthetic objects were then convolved with this PSF to simulate the blur. However, this approach is limited to scenes with known AR markers and uses a single, global PSF, which is unrealistic for spatially varying blur. More recent work \cite{prakash2025blind} addresses these limitations by leveraging off-the-shelf vision algorithms to separate and estimate defocus, motion, and sensor noise, followed by parameter optimization. However, their method models DoF as a  quadratic function of depth, whereas our method is derived from an optical model. Alternative methods, such as \cite{kan2012physically}, use physically-based DoF rendering to produce realistic lens blur but still depend on markers within the scene. In contrast, our method doesn't requires any prior knowledge of markers or camera parameters, making it robust across a wide range of camera setups. Other relevant works include \cite{mandl2021neural} and \cite{mandl2024neural}. The former estimates camera parameters using a multilayer perceptron (MLP) trained on camera-specific calibration images, such as those containing known markers and color charts. The latter, a follow-up by the same authors, focuses on synthesizing high-quality out-of-focus effects for MR applications. While both approaches rely on camera-specific training, our method does not require any prior calibration and can be applied to arbitrary real-world images or videos. 

\section{Problem Analysis}
\label{sec:analysis_sota}

\begin{table*}[htb] 
\caption{Methods used for evaluating the CoC map estimation and the reblurring module respectively. }
\label{tab:summarize_method}
\begin{center}
    \resizebox{0.85\linewidth}{!}{%
        \begin{tabular}{ccccccc}
        \toprule
           Motivation &   Input  & Operator & Method & RMSE $\downarrow$ & PSNR $\uparrow$ & SSIM $\uparrow$ \\
            \hline
             \multirow{2}{*}{Evaluate CoC map quality} & Defocus Img. & RGB \textrightarrow Disparity \textrightarrow CoC & Ranftl et al. \cite{ranftl2020towards} & $6.29$ & $-$ & $-$\\
              &Defocus Img. & RGB\textrightarrow CoC & Ruan et al. \cite{ruan2024self} & $4.07$ & $-$ &$-$ \\
             \hline
             \multirow{2}{*}{Evaluate reblur quality}  & Sharp Img. and CoC GT & Gaussian blur & Prakash et al. \cite{prakash2025blind}  & $-$ &$28.13$ & $0.90$\\
             & Sharp Img. and CoC GT &  \textcolor{black}{Network-based} blur &  Peng et al. \cite{peng2022bokehme} & $-$ &$31.08$ & $0.89$ \\
            \bottomrule
        \end{tabular}}
\end{center}
\end{table*}

\textbf{Motivation:} Compositing quality depends on both CoC map estimation and lens blur rendering. To better understand their individual impact, we review recent approaches for each component and conduct a systematic evaluation. Rather than focusing on full compositing pipelines, we isolate and assess individual methods using sythetic data. This provides access to ground-truth CoC maps, camera parameters, and scene depth, enabling controlled comparisons. We generate paired shallow DoF and all-in-focus images to evaluate lens blur performance under consistent conditions. This preliminary study helps identify optimal solutions for both CoC estimation and lens blur rendering.\\

\noindent \textbf{Prior solutions:} To assess the performance of CoC map estimation, we select two representative approaches. The first is a widely adopted method that begins by estimating disparity \cite{ranftl2020towards}, followed by computing the CoC map as a quadratic function of depth. The parameters of this function are trained and constrained using a Gaussian blur-based reblurring module, as described in \cite{prakash2025blind}. As an alternative, we include a recent method \cite{ruan2024self} that directly estimates the CoC map from RGB images. This approach has not been previously applied to compositing tasks involving lens blur. For evaluating reblurring methods, we focus on approaches that use ground-truth CoC maps and all-in-focus images as input. This setup allows us to evaluate on the effect of the reblurring process without interference from other intermediate results. We consider the standard Gaussian blur method \cite{prakash2025blind} and a neural network-based reblurring method \cite{peng2022bokehme} as representative examples. Although the method proposed by David et al.~\cite{mandl2021neural} is a strong neural network-based approach, we were unable to include it in our evaluation due to the unavailability of the code and pretrained weights.\\

\noindent \textbf{Experiment and findings:} Table~\ref{tab:summarize_method} summarizes the methods used in our evaluation. We tested each method on a single scene, as shown in Fig.~\ref{fig:spatially_varying_blur}, using nine defocused image variations generated by combining three different focus distances and three aperture sizes. Note that at this stage, our focus is on method evaluation and does not include the compositing task.

For CoC map estimation, we use the RMSE metric. Since CoC values estimated from depth or disparity are typically reported as the standard deviation ($\sigma$) of the Gaussian blur kernel, which is empirically set to one-fourth of the CoC diameter \cite{lee2019deep}, we multiply the output of these methods by a factor of four to report consistent CoC values. The results show that directly estimating the CoC map yields better performance compared to depth-based approaches. For reblurring methods, we evaluate performance using PSNR and SSIM, which indicate that the network-based approach achieves superior results. This can be attributed to the fact that the method in \cite{ranftl2020towards} was trained on sharp images; when applied directly to defocused inputs, its accuracy degrades. \textcolor{black}{In addition, it models the CoC as a quadratic function of disparity, constrained by Gaussian-based reblurring, which has been shown to be less accurate.} In contrast, the method proposed by \cite{ruan2024self} is specifically designed for defocused image input, resulting in improved CoC map quality. Regarding the reblurring module, the method in \cite{peng2022bokehme} benefits from training on a large-scale dataset, enabling it to produce more natural and visually realistic defocus blur.

The experiment results motivate us to directly estimate the CoC map from RGB images and to employ a neural network for lens blur rendering. This approach aligns with our goal of eliminating the need for depth information and camera metadata while maintaining high-quality performance, and it offers valuable insights for its application to the compositing task.\\

\noindent \textbf{Additional challenges for composites:} While direct estimation of the CoC map from RGB images (referred to as RGB-to-CoC) has demonstrated competitive performance, it poses specific additional challenges when applied to compositing tasks. The CoC values associated with virtual objects are not directly compatible with those derived from a captured photograph. This discrepancy arises because virtual objects, when inserted into a scene, occlude background regions and occupy different depth planes. For instance, as illustrated in Fig. \ref{fig:projection_mask}, the point on the virtual horse (orange point) cannot adopt the CoC value of the occluded background region (grey point), as these points exist at different depths in the scene. This issue is particularly relevant in our approach, which operates directly on the CoC map without access to complete scene depth information. Consequently, the challenge becomes how to reliably infer the correct CoC values for virtual objects during compositing.

\section{Method}
\label{sec:method}

To address the aforementioned challenges, this section revisits the relationship between the \textcolor{black}{CoC} and depth, based on the thin lens model, and presents the proposed solution in detail.

\begin{figure}
	\centering
	\includegraphics[width=1\linewidth]{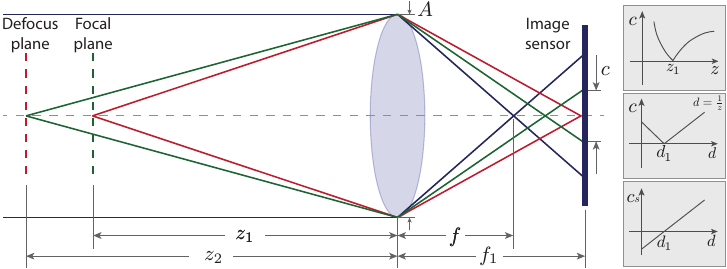}
	\caption{Illustration of Circle of Confusion \textcolor{black}{(CoC)} Calculation. The CoC is computed based on camera metadata, including focal length, focus distance, and sensor size. The right column illustrates the relationships between the defocus map and depth, the defocus map and disparity (the inverse of depth), and the signed defocus map and disparity. Our method leverages the linear relationship between the signed defocus map and disparity.}
	\label{fig:coc}
\end{figure}

\begin{figure*}[t]
    \centering
    \includegraphics[width=\linewidth]{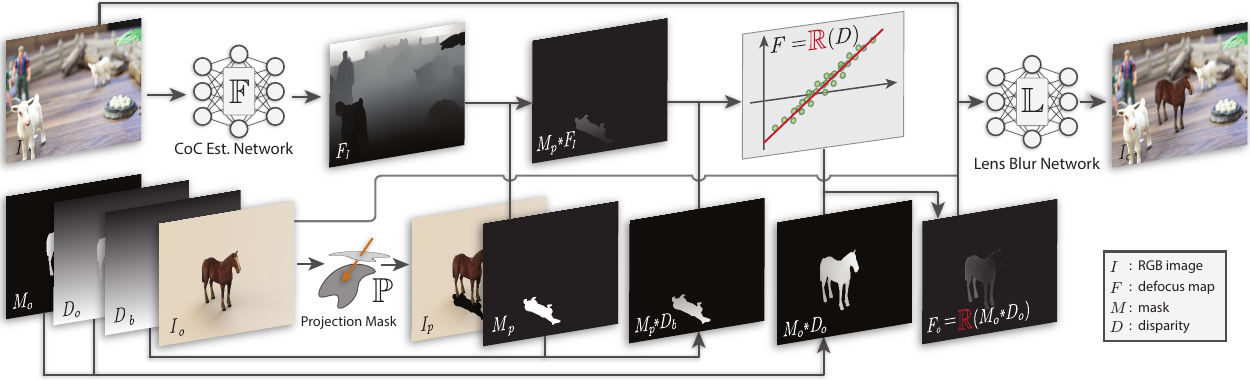}
    \caption{Our approach comprises a CoC map estimation, the linear fitting between the signed CoC in the real world and the disparity of virtual objects, as well as a neural reblurring network. The linear fitting is only applicable to a dedicated region, as defined by the projection mask. The CoC of the entire object is later inferred based on the fitted function. We adopt off-the-shelf methods for both CoC estimation and reblurring. Once the linear mapping is obtained, it can be readily applied to other virtual objects in the scene.}
    \label{fig:pipeline}
\end{figure*}

\subsection{Image formation model} 
\label{subsec:image_formation_model}
According to geometric optics (refer to Fig. \ref{fig:coc}), light rays emitted from points outside the focal plane (denoted as \(z_1\)) converge either in front of or behind the image sensor, resulting in a projected CoC with a diameter \(c\). In contrast, light rays originating from the focal plane are focused sharply on the sensor. Given a scene point at depth \(z\), along with the lens aperture diameter \(A\) and focal length \(f\), the diameter of the CoC \(c\) can be expressed following thin lens model \cite{potmesil1981lens}. Under the common assumption that the focal length \(f\) is significantly smaller than the focus distance \(z_1\), this expression for the CoC can be further simplified:

\begin{equation}
c = A\frac{\left|z_2-z_1\right|}{z_2}\frac{f}{z_1-f} \approx Af\frac{\left|z_2-z_1\right|}{z_2z_1} = Af\left|\frac{1}{z_1}-\frac{1}{z_2}\right|
\label{eq:coc}
\end{equation}

We demonstrate the relationship of depth $z$ and CoC size $c$ in Fig.~\ref{fig:coc} (right-top). Given that the CoC has a nonlinear relationship with depth, but a linear relationship with disparity, we can substitute disparity for depth to yield the simplified form \(c = A f |d - d_1|\) (Fig.~\ref{fig:coc} (right-middle)). Moreover, to achieve a linear relationship and simplify the equation, we can reparameterize it as a signed defocus formulation \cite{piche2023lens}, expressed as \(c_s = B(d - d_1)\) (Fig.~\ref{fig:coc} (right-bottom)), where $B$ is the scaled blur parameter and $d_1$ is the focus disparity. In this way, the relationship between the CoC and disparity is reformulated, which provides the key insight for our approach.

\begin{figure}
    \centering
    \includegraphics[width=\linewidth]{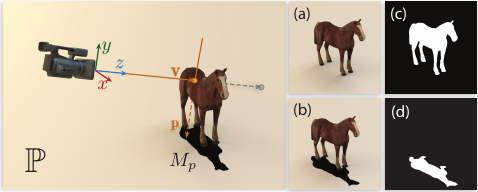}
    \caption{The projection mask of our approach. Left: we cast shadow along $\mathbf{p}$, where $\mathbf{p}\parallel -y$ and $\mathbf{p}\bot \mathbf{v}$, $\mathbf{v}$ the look-at vector of camera along $z$ direction. Right (camera view): (a) original object image. (b) cast shadow along $\mathbf{p}$. (c) original object mask. (d) our corrected projection mask.}
    \label{fig:projection_mask}
\end{figure}

\subsection{Neural lens}
\label{subsec:our_approach}

As illustrated in Figure~\ref{fig:pipeline}, our method consists of few key components: a CoC map estimation network $\mathbb{F}$, a linear regression module $\mathbb{R}$ that establishes the relationship between the signed CoC in the real scene and the disparity of virtual objects, a projection mask rendering module $\mathbb{P}$, and a neural reblurring network $\mathbb{L}$ that synthesizes the final defocused image. We detailed each step shown below: \\

\vspace{-0.3cm}

\noindent\textbf{CoC Map:} We leverage off-the-shelf, high-quality CoC estimation methods that operate directly on RGB images. These methods are based on thin-lens modeling, and thus produce per-pixel CoC values that closely adhere to the theoretical CoC formulation. Among them, we adopt the learning-based approach proposed by Ruan et al. \cite{ruan2024self}, which aligns well with our requirements. However, since this method outputs only the absolute CoC values, we convert them to signed values to support our analysis. The defocus map is estimated by $F_I = \mathbb{F}(I)$.

For virtual objects rendering, we use a 360$^\circ$ camera (Insta360\footnote{Insta360: \url{https://www.insta360.com}}) to capture multi-exposure 360$^{\circ}$ images to have a high dynamic range environment map, allowing us to extract real-world illumination.
Shadows are rendered following the approach of \cite{rhee2017mr360}, utilizing differential rendering techniques \cite{debevec1998rendering}. As our primary focus is achieving high-fidelity composites with realistic DoF effects, global illumination, reflection, and refraction are not emphasized in this work, though they can be easily integrated using existing frameworks as mentioned in \cite{mandl2021neural}.

\noindent\textbf{Analysis:} The core of our approach lies in bridging the virtual and real worlds by leveraging the linear relationship between object disparity and the signed CoC of the scene. We start with an estimated CoC map from the real scene and the known depth of virtual objects, establishing correspondences through matched spatial positions. However, this point-to-point correspondence fails within object mask regions due to occlusion, as illustrated in Fig.\ref{fig:projection_mask} (left). When a ray is cast along the viewing direction $\mathbf{v}$, it intersects the foreground surface (orange point), providing the correct depth. In contrast, the estimated CoC corresponds to the background surface (gray point) behind the object, where $\mathbf{v}$ continues beyond the occlusion.
To address this mismatch, we introduce a projection mask rendering module $\mathbb{P}$, defining a spatial region where the real and virtual scenes are reliably aligned. Outside this region, correspondence breaks down due to occlusions or depth discontinuities. Given that only the CoC from the real scene and the depth from the virtual scene are available, we perform a linear regression solely within this valid region. The projection mask $M_p$ is rendered by casting rays along the direction $\mathbf{p}$, where $\mathbf{p} \parallel -y$ and $\mathbf{p} \perp \mathbf{v}$. The resulting shadow projected onto the background surface identifies the valid region (Fig.~\ref{fig:projection_mask} (d)) in the defocus map, compared to the original inaccurate mask (Fig.~\ref{fig:projection_mask} (c)).
\begin{figure*}[t]
    \centering
    \includegraphics[width=\linewidth]{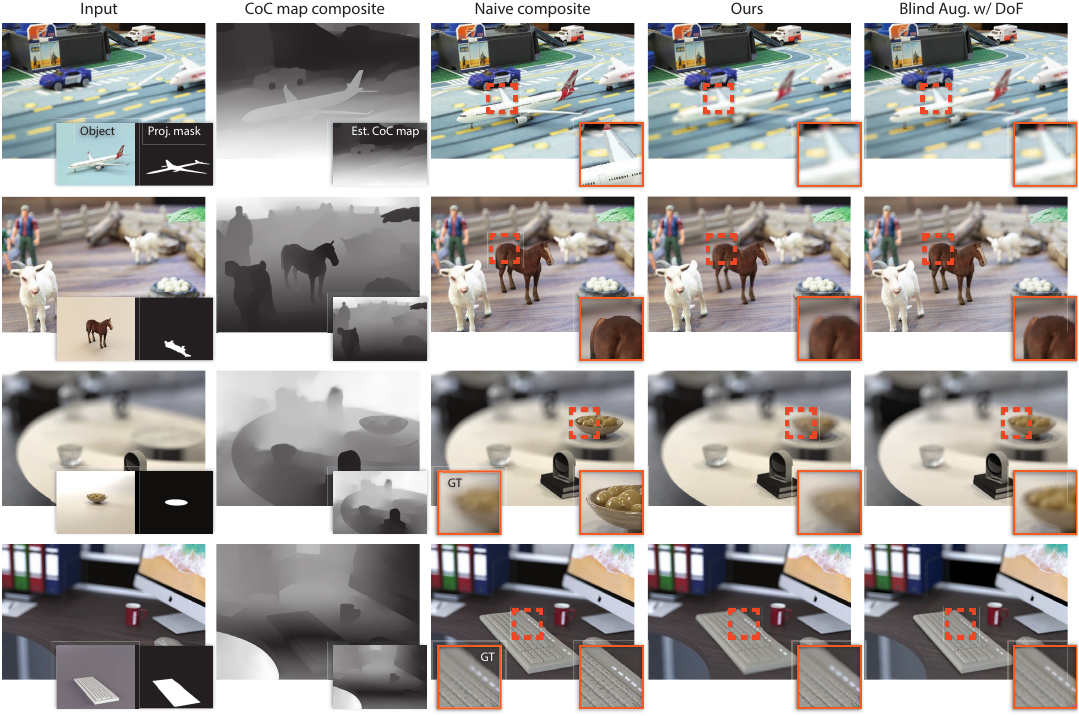}
    \caption{Please zoom in to better observe the compositing quality. Comparison between our method and Blind Augmentation \cite{prakash2025blind} with DoF effect. The first column shows the background photographs to be augmented. The second column presents the estimated CoC maps from \cite{prakash2025blind}, while the third shows the CoC maps used for reblurring virtual objects based on our proposed fitting scheme. The fourth and fifth columns display our compositing results and those from Blind Augmentation, respectively. The top two examples are real photographs, and the bottom two are rendered scenes with available ground truth, denoted as \textbf{GT}. Our method demonstrates superior performance, particularly around object boundaries (e.g., airplane, horse, bowl) and in maintaining consistent blur levels (e.g., horse, bowl, keyboard). See Sec.~\ref{subsec:comparison} for detailed discussion.}
    \label{fig:results}
\end{figure*}

To enable linear fitting, we convert the CoC map into a signed CoC representation. This conversion depends on the distribution of CoC values within the projection-masked region. If the CoC values do not include small or near-zero values, we assume a consistent defocus trend, characterized by a monotonic increase or decrease. This allows for a straightforward linear fit. In contrast, if small CoC values are present, indicating the presence of in-focus regions, we infer that the farthest depths correspond to negative values on the CoC axis. This assumption is based on the typical camera orientation where the viewing direction points downward toward the background surface. Accordingly, we shift the CoC map to a signed representation by assigning negative values to these far-background regions, thereby ensuring accurate modeling of defocus behavior across different depth layers. The linear relationship $\mathbb{R}$ is estimated by minimizing the mean square error:
\begin{equation}
\min_{a, b} \sum_{i=1}^n \left( F_i - (a D_i + b) \right)^2,
\end{equation}
where $F_i\in M_p\odot F_I$ and $D_i\in M_p\odot D_b$ denote the observed CoC size and disparity in pair, and $a, b$ are the parameters to be estimated. The fitted linear model between the signed CoC and disparity not only captures their relationship within the masked regions, but can also be readily applied to other video \textcolor{black}{frames (see Fig. \ref{fig:flying_plane_video}) and to other virtual objects (see Fig. \ref{fig:two_cars})}.\\

\noindent \textbf{Lens Reblur:} In our lens reblur module, we adopt \textcolor{black}{a neural network-based lens blur model $\mathbb{L}$ from \cite{peng2022bokehme}, which integrates classical kernel-scattering rendering with neural rendering. Existing compositing methods including recent learning-based approaches~\cite{mandl2021neural,mandl2024neural}, typically rely on traditional alpha blending to combine reblurred foreground and background layers, often leading to visible boundary artifacts. In contrast, our approach reformulates the input representation, embedding the compositing operation directly into the network itself.} We begin with naïve composites, where the virtual object remains sharp while the background image is naturally defocused. To guide the reblurring, we construct a CoC map by assigning appropriate CoC values to the virtual object, while setting the CoC to zero in all other regions. This ensures that the real photograph remains unchanged. By integrating the compositing task into the reblurring problem, we fully leverage the network’s strength in producing high-quality blur transitions at depth boundaries, which aligns with the composite object boundaries. The final image can be composited as:
\begin{equation}
    I_c = \mathbb{L}(\mathbb{R}(M_o \odot D_o), I, I_o),
\end{equation}
where $M_o$, $D_o$, and $I_o$ are the object's mask, depth, and RGB image respectively. 
As illustrated in Fig.~\ref{fig:user_study_result} and Fig.~\ref{fig:results} , this strategy achieves visually coherent and high-quality results.

\section{Results}
We demonstrate the effectiveness of our method in naturally augmenting real-world scenes with virtual objects and compare it against state-of-the-art approaches (Sec. \ref{subsec:comparison} both on the real world data and the synthetic data). A user study evaluates perceptual composition quality (Sec. \ref{sec:user_study}).

\subsection{Comparison}
\label{subsec:comparison}
We compare our method with Blind Augmentation \cite{prakash2025blind}, which jointly addresses depth-of-field (DoF), motion blur, and noise in a blind manner. Although it shares a similar motivation with our approach in avoiding the reliance on camera metadata, our evaluation specifically focuses on the DoF component. To ensure a fair comparison, we isolate their DoF module by disabling the motion blur and noise effects. Following their pipeline, we incorporate the defocus deblurring technique from \cite{ruan2022learning} and utilize the disparity estimation method from \cite{ranftl2020towards} to obtain the necessary depth information for their method. Other related approaches \cite{mandl2021neural, mandl2024neural} require prior camera calibration and access to metadata, and their implementations are not publicly available to date, therefore, they are not included in our comparison.
The qualitative (Sec.\ref{sub:qualtitative})  and quantitative (Sec.\ref{sub:quantitative}) results presented below demonstrate that our method outperforms the Blind Augmentation approach, particularly in handling compositing boundaries and achieving blur level more consistently.
\begin{figure}
    \centering
    \includegraphics[width=\linewidth]{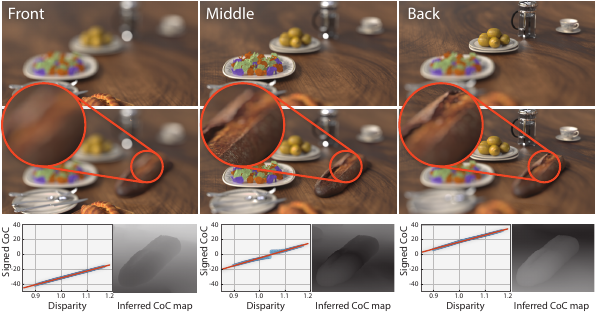}
    \caption{Results of object insertion under different defocus levels: focusing on the front (croissant), the middle (baguette), and the back (coffee plunger). The first row shows the original background images prepared for compositing. The second row presents our reblurred results based on the inferred CoC maps. The bottom row illustrates our fitted relationship between the background signed CoC and the disparity of the composited object (baguette in this case). Our method produces fine-grained and accurate CoC maps for the inserted object, enabling realistic and natural compositing.  Darker regions indicate areas in focus, while brighter regions represent areas that are increasingly out of focus.}
    \label{fig:spatially_varying_blur}
\end{figure}
\subsubsection{Qualitative}
\label{sub:qualtitative}

Figure~\ref{fig:results} presents a comparison between our method and the Blind Augmentation approach proposed by \cite{prakash2025blind}, using a diverse set of virtual objects. The first two rows illustrate composites generated with real photographs, while the bottom two rows display fully rendered scenes where ground truth is available for reference. In the first row, a virtual horse is composited with spatially varying blur, where the head remains in focus and the tail is defocused. The second row shows a virtual airplane positioned on a runway. Other examples include composited virtual keyboards and bowls. The qualitative results demonstrate that our method produces accurate CoC maps for virtual objects. This effectiveness is particularly evident, as the estimated CoC maps are used to guide reblurring, resulting in an appearance that closely matches the ground truth, for example, the bowl and keyboard shown in the figure.

\begin{figure}[t]
    \centering
    \includegraphics[width=\linewidth]{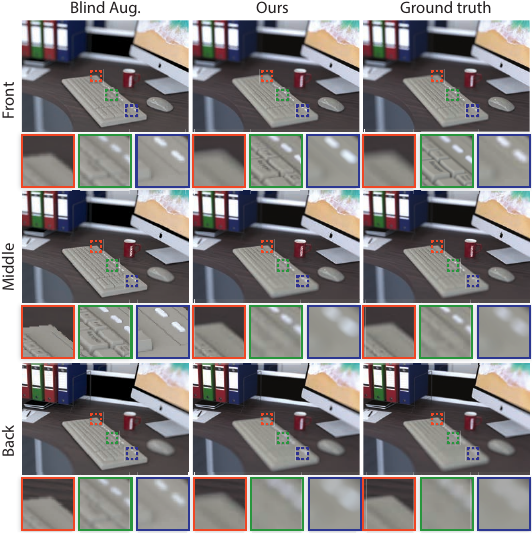}
    \caption{Our approach produces highly realistic spatially varying defocus blur for the same inserted object across different focus distances, closely matching the ground truth and outperforming Blind Augmentation. The synthetic scene is rendered to simulate real-world scenarios with a "virtual" keyboard inserted, specifically for quantitative evaluation with ground truth available. Input images are omitted here for brevity; one of the inputs can be referenced in the bottom row of Fig.~\ref{fig:results}.}
    \label{fig:comparison_spatially_varying}
\end{figure}

Figure~\ref{fig:spatially_varying_blur} further demonstrates the effectiveness of our approach in handling spatially varying blur and resolving scale ambiguity between real and virtual scenes. The fitted curves in the figure represent three focus scenarios: foreground (croissant), midground (baguette), and background (coffee plunger). These results confirm that our model captures the relationship between depth and CoC in a physically meaningful way, ensuring realistic and consistent blur effects even without access to camera metadata. In addition to this,   Fig.~\ref{fig:comparison_spatially_varying} presents a comparison between our method and Blind Augmentation in the task of compositing a keyboard object. The results indicate that our approach exhibits superior robustness and consistency with respect to the ground truth, particularly for virtual objects exhibiting varying levels of blur.

In contrast to our method, the Blind Augmentation approach \cite{prakash2025blind} models depth-of-field effects through a two-stage process involving deblurring followed by reblurring. Specifically, it first applies a defocus deblurring method \cite{ruan2022learning}, then performs Gaussian reblurring guided by a quadratic function fitted between the CoC and disparity. In this framework, CoC values for virtual objects are inferred from disparity, based on the assumption that the real and virtual scenes share the same scale. However, this assumption often fails in practical scenarios, as the method does not explicitly account for scale mismatches between domains.
While Blind Augmentation performs reasonably well when virtual objects are inserted at discrete depth levels, its performance degrades when compositing objects with spatially varying blur, such as the keyboard and horse examples shown in Fig.~\ref{fig:results}. In contrast, our method produces more visually coherent results by leveraging a physically meaningful mapping between CoC and disparity across the real and virtual domains. This formulation makes our approach inherently more robust to depth variations and scale ambiguity between real and virtual content.

Another key advantage of our approach lies in its handling of compositing boundaries. Prior works typically rely on alpha blending techniques \cite{mandl2021neural, mandl2024neural, prakash2025blind}. For instance, Blind Augmentation applies erosion and dilation near depth discontinuities to smooth transitions. However, when the blur is strong and there is high color contrast between the object and the background, these methods often produce noticeable artifacts, as illustrated in Fig.~\ref{fig:results}. We address this issue by leveraging the properties of a reblurring neural network, as discussed in Sec.~\ref{subsec:our_approach}. Rather than using a fully sharp input as originally intended, we feed a naively composited image into the network and modify the CoC map such that only the virtual object contains valid values, while the background is set to zero. This adaptation allows us to exploit the strengths of the existing reblurring network while avoiding the boundary artifacts that commonly affect traditional compositing methods. The insets in Fig.~\ref{fig:results} show that the Blind Augmentation method exhibits obvious boundary artifacts, whereas our approach achieves significantly improved visual quality.

\begin{figure}
    \centering
    \includegraphics[width=\linewidth]{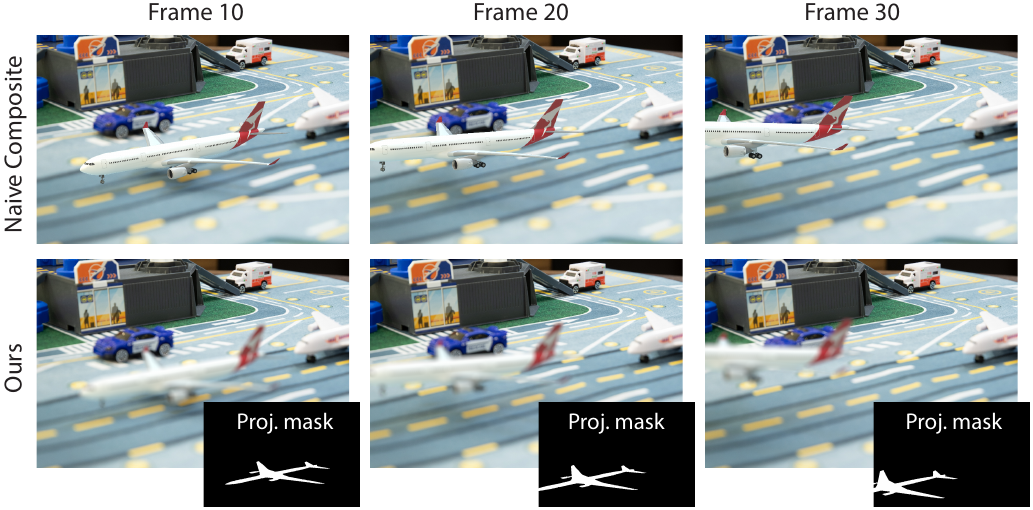}
    \caption{\textcolor{black}{Selected frames illustrating our compositing results for an airplane takeoff sequence. Our method produces temporally consistent lens blur across video frames. Note that the regression model is derived from the first frame using the projected mask region. For the remaining frames, only the virtual object disparity is required, and their CoC values are predicted using the same fitted regression model. The projection masks shown in those frames are for visualization purposes only. Please refer to the supplementary video for the full animation.}}
    \label{fig:flying_plane_video}
\end{figure}

\begin{figure}[t]
    \centering
    \includegraphics[width=\linewidth]{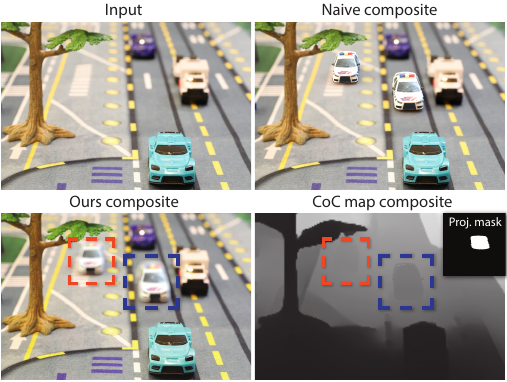}
    \caption{Our approach extends naturally to multiple objects using a shared linear relationship between object disparity and the signed CoC of the photographic scene. The car in the blue frame with its projected mask is used to fit the mapping between the real and virtual worlds. The CoC of the car in red frame is then inferred directly from this model, resulting in a larger CoC (brighter region) and stronger blur, consistent with the depth and appearance of the real world case.}
    \label{fig:two_cars}
\end{figure}

Once the linear fitting is performed, the resulting model can be generalized and reused for multiple virtual objects, as our method establishes a linear mapping between object disparity and the signed CoC. This enables consistent blur rendering across different virtual objects. As demonstrated with the two cars in Fig.~\ref{fig:two_cars}, the mapping between the real and virtual domains is derived using the right car, while the CoC value for the left car is directly inferred using the fitted model. The left car, positioned farther from the focus (sharp) region, exhibits a larger CoC value and appears brighter in the composited CoC map. This outcome is physically reasonable and aligns with real-world depth-of-field behavior. These findings confirm that our approach supports consistent and spatially coherent rendering of multiple virtual objects. Moreover, the fitted model can be readily extended to support video compositing tasks. \textcolor{black}{As illustrated in Fig.~\ref{fig:flying_plane_video}, we present three representative frames depicting a virtual plane taking off, demonstrating the temporal coherence of lens blur produced by our compositing method. The regression model is fitted using only the first frame, where the projected mask region is required. For subsequent frames, the model can directly predict the CoC for reblurring. Please refer to the supplementary video for the complete sequence, as well as an additional example featuring a slithering snake. \\}

\subsubsection{Quantitative}
\label{sub:quantitative}
In addition to qualitatively evaluating our method’s ability to consistently composite virtual objects into real-world scenes, we further conduct a quantitative assessment. The dataset is synthetically generated in Blender using the Cycles ray-tracing engine to simulate realistic DoF effects. To mimic the real compositing process, we render: (1) a defocused background image without the virtual object, (2) the virtual object alone, and (3) the ground-truth composite image where the object is correctly blended into the background. This enables precise quantitative evaluation of the compositing quality.
\\

\begin{table}[t] 
\caption{The quantitative evaluation of composite quality was conducted on our rendered scenes. To better reflect the performance, we report results only within the composite regions defined by bounding boxes across all test cases.}
\label{tab:quantitative_result}
\small
\begin{center}
    \resizebox{0.9\linewidth}{!}{
        \begin{tabular}{p{2.5cm}>{\centering\arraybackslash}p{1.5cm}>{\centering\arraybackslash}p{1.5cm}}
        \toprule
           Method &   PSNR $\uparrow$  & SSIM $\uparrow$ \\
            \hline
             Naive Composites  & 26.41 & 0.73 \\
              Blind Aug. w/ DoF & 27.88 & 0.82 \\
             Ours & \textbf{34.66} & \textbf{0.97}   \\
            \bottomrule
        \end{tabular}}
\end{center}
\end{table}

\noindent\textbf{Data:} We used three distinct 3D scenes, each with a different focus distance. This setup allows virtual objects to appear with varying levels of blur, enabling evaluation of the robustness of our approach under diverse depth-of-field (DoF) conditions. In total, we generated nine variations across the three scenes for evaluation. Quantitative results were reported by computing PSNR and SSIM between the compositing results and the rendered ground truth.
\\
\begin{table}[t] 
\caption{The result of user study. P-values from statistical comparisons with Ours are reported, with significant results ($p < 0.01$) highlighted in bold. Note that we adopt blind augmentation \cite{prakash2025blind} using only the depth-of-field (DoF) effect.}
\label{tab:user_study}
\small
\begin{center}
    \resizebox{0.9\linewidth}{!}{%
        \begin{tabular}{p{2.5cm}>{\centering\arraybackslash}p{1.5cm}>{\centering\arraybackslash}p{1.5cm}}
        \toprule
           Method &   Failure Rate $\uparrow$  & p =  \\
            \hline
             Naive Composites  & 74.9\% & \textbf{1.9954e-13} \\
              Blind Aug. w/ DoF & 75.1\% & \textbf{9.5198e-15} \\
             Ours & \textbf{90}\% & -   \\
            \bottomrule
        \end{tabular}}
\end{center}
\end{table}

\noindent \textbf{Result:} Table~\ref{tab:quantitative_result} presents the quantitative results comparing our method with the baseline. Since the background photograph remains unchanged except for the added virtual object, we focus the evaluation on the composited area. Specifically, we first obtain the virtual object region using its mask and then compute the tightest bounding box enclosing this region. PSNR and SSIM are calculated within this bounding box, to ensure that the evaluation specifically reflects the quality of the composited content and avoids including irrelevant background regions.

We use PSNR and SSIM to assess the perceptual quality of the RGB composites, where PSNR reflects pixel-wise reconstruction fidelity, and SSIM captures structural similarity and perceptual consistency. The results show that our method produces outputs closer to the ground truth, achieving a PSNR improvement of 6.78 dB and an SSIM increase of 0.15 over Blind Augmentation. This improvement is largely due to Blind Augmentation's tendency to introduce artifacts and produce less consistent blur when handling spatially varying blur across the object surface, as demonstrated in Fig.\ref{fig:results} and Fig.\ref{fig:comparison_spatially_varying}.

\begin{figure}[t]
    \centering
    \includegraphics[width=\linewidth]{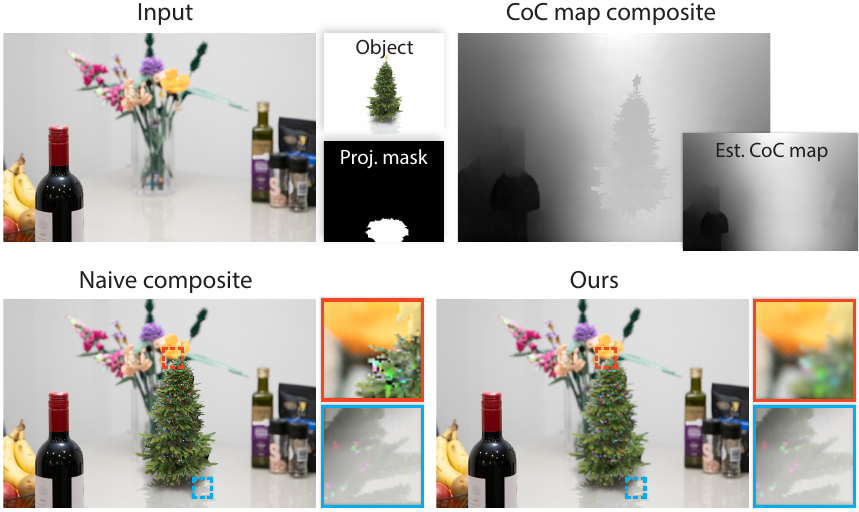}
    \caption{\textcolor{black}{We evaluate our method on a challenging tabletop scene with reflective surfaces (tempered glass), textureless regions (wall and table), and complex geometry (LEGO bouquet), augmented with a virtual Christmas tree decorated with fairy lights. Despite difficulties in accurate CoC estimation, our method achieves visually plausible lens blur effects, as shown in the red inset, primarily because only the CoC values within the projected mask region influence the final compositing. A limitation of our current method lies in handling lens blur for shadows on reflective surfaces, as highlighted in the blue inset where the light shadow should appear blurred accordingly. Zoom in for better visualization. }}
    \label{fig:limitation}
\end{figure}

\begin{figure*}[t]
    \centering
    \includegraphics[width=\linewidth]{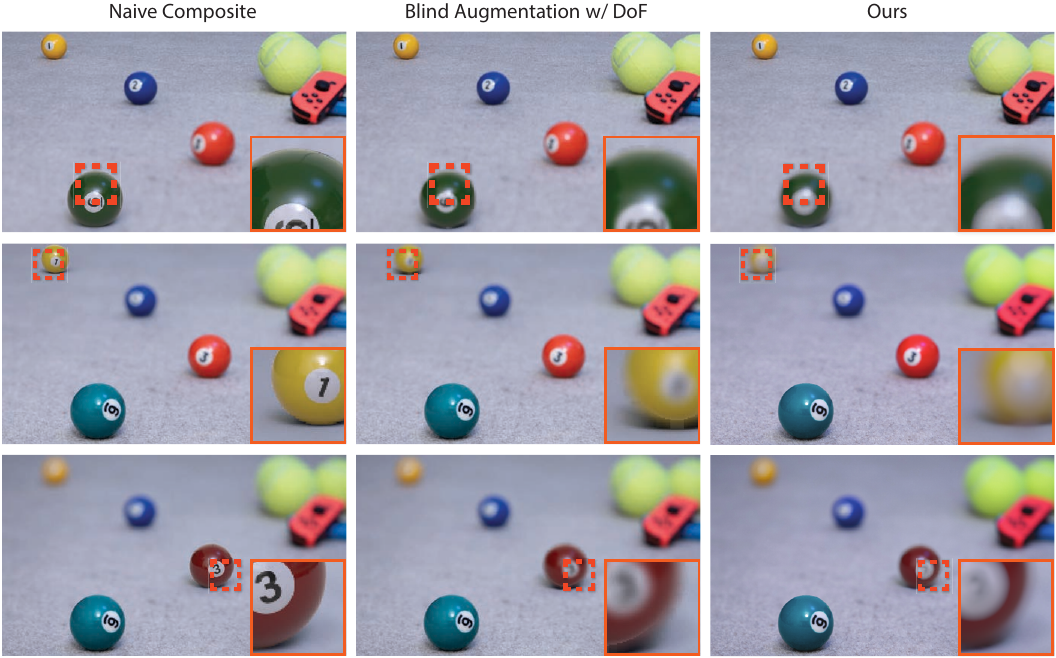}
    \caption{Comparison between different compositing methods used in the user study. Red, green, and yellow virtual billiard balls are alternately composited into the scene, each undergoing three levels of blur. Rows correspond to different ball colors: green (top), yellow (middle), and red (bottom). All synthetic objects share the same shape to eliminate bias from texture differences. We compare our method with naïve compositing and blind augmentation. The DoF effect is included in the blind augmentation only for comparison. Neither baseline method uses calibration or camera metadata. Blind augmentation often introduces boundary artifacts due to erosion operations and shows inconsistent blur levels due to scale ambiguity between virtual and real-world elements. Our method addresses both issues and achieves more consistent and realistic results.}
    \label{fig:user_study_result}
\end{figure*}

\subsection{User Study}
\label{sec:user_study}

We conducted a user study to evaluate the visual fidelity of our method. The study involved human participants, and all ethical and experimental procedures were approved by the research ethics and integrity department of the authors' institution.
\\

\noindent\textbf{Stimuli and Task:} The stimuli consisted of still images incorporating DoF blur effects with augmentations, similar to those shown in Fig.~\ref{fig:user_study_result}. Each image included a single virtual object (a billiard ball) composited into a photographed scene with real billiard balls using three methods: naive compositing, blind augmentation \cite{prakash2025blind}, and our proposed approach. As described in Sec. \ref{subsec:comparison}, we focus exclusively on DoF blur effects, with motion blur and noise disabled. Billiard balls were chosen as the virtual object to maintain consistent shape and appearance across all images, thereby preventing participants from identifying virtual content based on unique object geometry or texture. This ensures that judgments are based solely on the naturalness and consistency of the DoF effect.

The virtual object was placed in three different spatial positions. Each position was composited into a set of nine real photographs with variations in aperture and focus distance (three settings each). Across three rendering methods, this resulted in a total of 81 stimuli (3 positions × 3 apertures × 3 focus distances × 3 methods). All the images are randomized and the user will be viewed per image per time. They were asked: ``Do you believe there are virtual billiard balls in the image? If yes, please click the specific colour of the ball, Otherwise, select No". Users are told it might have one, multiple, or no virtual objects to encourage user focus on their visual quality. Our goal is to access the quality of DoF effect on the objects in various blur conditions when blending virtual objects into real photograph.
\\

\noindent\textbf{Outcome:} The results of the user study are presented in Table \ref{tab:user_study}. We recruited a total of 25 participants, with ages ranging from 21 to 34. Participants’ prior knowledge of the task ranged from “no knowledge” to “familiar” with computer graphics. The participants were compensated for their contribution. We report the percentage of trials in which participants were unable to identify the virtual objects. A higher percentage indicates better perceptual realism, following the evaluation protocol proposed in \cite{prakash2025blind}. To assess statistical significance, we conducted a two-proportion z-test \cite{building2023}, comparing our method against the baseline approaches. The results show that our method significantly outperforms the other two methods, with very small p-values indicating strong statistical significance.

\section{Discussion / Limitations}
\textcolor{black}{
The proposed method enables practical and flexible scene compositing without the need for prior camera metadata, scene depth, or calibration, and it demonstrates competitive performance. However, it presents several limitations, particularly in handling reflective surfaces and achieving real-time computational efficiency.\\
\indent We evaluate our approach in a challenging real-world scenario, as depicted in Fig.~\ref{fig:limitation}, which features a reflective tempered-glass tabletop, textureless regions such as walls and glass surfaces, and a LEGO bouquet with complex geometry. A virtual Christmas tree decorated with fairy lights is composited into the scene and placed close to the LEGO bouquet, resulting in partial occlusion. This setup introduces extra challenges for accurate scene understanding. \\
\indent A key difficulty lies in estimating the CoC in textureless and reflective regions. In this scenario, the glass tabletop and walls exhibit similarly low-texture and low-contrast characteristics, which hinder accurate CoC inference. These limitations arise primarily from the dependency on neural network-based components for CoC estimation, whose performance is inherently constrained by the scale and diversity of the training data, particularly due to the limited representation of reflective surfaces. However, our method is still able to produce reasonable lens blur and bokeh effects for the virtual tree, as shown in Fig.~\ref{fig:limitation}. This is mainly because the blur synthesis of virtual objects depends solely on the projected mask region, which, in this case, is estimated with sufficient accuracy.\\
\indent Another limitation involves the rendering of cast shadows. Our method currently applies differential rendering~\cite{debevec1998rendering} directly to generate shadows, treating them as part of the photographic background without separate lens blur processing. While this approach works effectively on matte surfaces, as illustrated in most of our examples, it results in visually unrealistic sharp shadows when applied to reflective materials. Specifically, shadows cast by virtual fairy lights onto the tempered-glass table remain unnaturally sharp, whereas they should appear blurred. This discrepancy reduces visual realism and remains an open issue not adequately addressed in existing literature, warranting future research.\\
\indent Regarding runtime performance, our current implementation prioritizes compositing quality over computational efficiency. The CoC estimation takes approximately one second per photograph and the lens blur rendering requires around 180 milliseconds per frame using an NVIDIA RTX 8000 GPU at a resolution of 512 × 960 pixels. 
Since CoC needs to be estimated only once per photograph, for example in our animation videos featuring a plane taking off or a snake slithering, the overall rendering speed is primarily determined by the lens blur module. Although the current implementation does not yet achieve real-time performance, it has the potential to reach real-time speed with slight compromises in visual quality, such as reducing the lens blur network size. Future work will address computational efficiency, as both the CoC estimation network and the lens blur network can be improved through model compression strategies such as quantization and pruning~\cite{liu2021discrimination}. These techniques have the potential to enable real-time deployment on resource-constrained platforms, including integration with Meta Quest using the Unity Sentis framework\footnote{Unity Sentis: \url{https://unity.com/products/sentis}}.}

\section{Conclusion}
We have presented a novel approach for augmenting real-world photographs with virtual objects in a visually consistent manner under lens blur, without requiring prior calibration, scene depth information, or camera metadata. The core contribution of our method lies in bridging the real and virtual domains by exploiting the linear relationship between object disparity and the signed Circle of Confusion derived from the real image. This formulation effectively resolves the scale ambiguity inherent in prior methods when the scale of virtual objects is not aligned with that of the real scene. To the best of our knowledge, this is the first work to directly utilize per-pixel CoC instead of relying on disparity, offering an alternative method for compositing in augmented reality applications.

We conducted comprehensive comparisons with recent state-of-the-art technique, demonstrating that our approach delivers superior results, particularly in scenarios where virtual objects exhibit varying levels of blur. Our method achieves improved blur consistency and more seamless integration at compositing boundaries. In addition to quantitative and qualitative evaluations, a user study further confirms that our method significantly outperforms existing alternative in terms of visual fidelity.\\
\vspace{-0.1cm}
\section{Acknowledgements}
This work was supported by the Culture, Sports and Tourism R\&D Program through the Korea Creative Content Agency grant funded by the Ministry of Culture, Sports and Tourism, South Korea (Project Number: RS-2024-00399136).

\bibliographystyle{abbrv-doi}

\bibliography{template}

@inproceedings{ruan2024self,
  title={Self-supervised video defocus deblurring with atlas learning},
  author={Ruan, Lingyan and B{\'a}lint, Martin and Bemana, Mojtaba and Wolski, Krzysztof and Seidel, Hans-Peter and Myszkowski, Karol and Chen, Bin},
  booktitle={Proceedings of the ACM SIGGRAPH},
  pages={1--11},
  year={2024}
}

@article{prakash2025blind,
  title={Blind Augmentation: calibration-free camera distortion model estimation for real-time mixed-reality consistency},
  author={Prakash, Siddhant and Walton, David R and Anjos, RKD and Steed, Anthony and Ritschel, Tobias},
  journal={IEEE Transactions on Visualization and Computer Graphics},
  year={2025},
}

@inproceedings{mandl2021neural,
  title={Neural Cameras: learning camera characteristics for coherent mixed reality rendering},
  author={Mandl, David and Roth, Peter M and Langlotz, Tobias and Ebner, Christoph and Mori, Shohei and Zollmann, Stefanie and Mohr, Peter and Kalkofen, Denis},
  booktitle={Proceedings of IEEE International Symposium on Mixed and Augmented Reality},
  pages={508--516},
  year={2021},
  organization={IEEE}
}

@inproceedings{mandl2024neural,
  title={Neural Bokeh: learning lens blur for computational videography and out-of-focus mixed reality},
  author={Mandl, David and Mori, Shohei and Mohr, Peter and Peng, Yifan and Langlotz, Tobias and Schmalstieg, Dieter and Kalkofen, Denis},
  booktitle={Proceedings of IEEE Conference on Virtual Reality and 3D User Interfaces},
  pages={870--880},
  year={2024},
  organization={IEEE}
}

@inproceedings{lee2019deep,
  title={Deep defocus map estimation using domain adaptation},
  author={Lee, Junyong and Lee, Sungkil and Cho, Sunghyun and Lee, Seungyong},
  booktitle={Proceedings of the IEEE/CVF Conference on Computer Vision and Pattern Recognition},
  pages={12222--12230},
  year={2019}
}

@inproceedings{piche2023lens,
  title={Lens Parameter Estimation for Realistic Depth of Field Modeling},
  author={Pich{\'e}-Meunier, Dominique and Hold-Geoffroy, Yannick and Zhang, Jianming and Lalonde, Jean-Fran{\c{c}}ois},
  booktitle={Proceedings of the IEEE/CVF International Conference on Computer Vision},
  pages={499--508},
  year={2023}
}

@article{wang2018deeplens,
  title={DeepLens: shallow depth of field from a single image},
  author={Wang, Lijun and Shen, Xiaohui and Zhang, Jianming and Wang, Oliver and Lin, Zhe and Hsieh, Chih-Yao and Kong, Sarah and Lu, Huchuan},
  journal={ACM Transactions on Graphics},
  volume={37},
  number={6},
  pages={1--11},
  year={2018},
  publisher={ACM New York, NY, USA}
}

@inproceedings{peng2022bokehme,
  title={Bokehme: When neural rendering meets classical rendering},
  author={Peng, Juewen and Cao, Zhiguo and Luo, Xianrui and Lu, Hao and Xian, Ke and Zhang, Jianming},
  booktitle={Proceedings of the IEEE/CVF Conference on Computer Vision and Pattern Recognition},
  pages={16283--16292},
  year={2022}
}

@article{potmesil1981lens,
  title={A lens and aperture camera model for synthetic image generation},
  author={Potmesil, Michael and Chakravarty, Indranil},
  journal={ACM SIGGRAPH Computer Graphics},
  volume={15},
  number={3},
  pages={297--305},
  year={1981},
  publisher={ACM New York, NY, USA}
}

@article{ranftl2020towards,
  title={Towards robust monocular depth estimation: Mixing datasets for zero-shot cross-dataset transfer},
  author={Ranftl, Ren{\'e} and Lasinger, Katrin and Hafner, David and Schindler, Konrad and Koltun, Vladlen},
  journal={IEEE Transactions on Pattern Analysis and Machine Intelligence},
  volume={44},
  number={3},
  pages={1623--1637},
  year={2020},
  publisher={IEEE}
}

@InProceedings{Vitor2023towards,
    author    = {Guizilini, Vitor and Vasiljevic, Igor and Chen, Dian and Ambruș, Rareș and Gaidon, Adrien},
    title     = {Towards Zero-Shot Scale-Aware Monocular Depth Estimation},
    booktitle = {Proceedings of the IEEE/CVF International Conference on Computer Vision},
    
    year      = {2023},
    pages     = {9233-9243}
}

@article{zhang2021joint,
  title={Joint depth and defocus estimation from a single image using physical consistency},
  author={Zhang, Anmei and Sun, Jian},
  journal={IEEE Transactions on Image Processing},
  volume={30},
  pages={3419--3433},
  year={2021},
  publisher={IEEE}
}

@article{ruan2021aifnet,
  title={AIFNet: All-in-focus image restoration network using a light field-based dataset},
  author={Ruan, Lingyan and Chen, Bin and Li, Jizhou and Lam, Miuling},
  journal={IEEE Transactions on Computational Imaging},
  volume={7},
  pages={675--688},
  year={2021},
  publisher={IEEE}
}

@inproceedings{gur2019single,
  title={Single image depth estimation trained via depth from defocus cues},
  author={Gur, Shir and Wolf, Lior},
  booktitle={Proceedings of the IEEE/CVF Conference on Computer Vision and Pattern Recognition},
  pages={7683--7692},
  year={2019}
}

@article{lee2010real,
  title={Real-time lens blur effects and focus control},
  author={Lee, Sungkil and Eisemann, Elmar and Seidel, Hans-Peter},
  journal={ACM Transactions on Graphics},
  volume={29},
  number={4},
  pages={1--7},
  year={2010},
  publisher={ACM New York, NY, USA}
}

@article{yu2010real,
  title={Real-time depth of field rendering via dynamic light field generation and filtering},
  author={Yu, Xuan and Wang, Rui and Yu, Jingyi},
  journal={Computer Graphics Forum},
  volume={29},
  number={7},
  pages={2099--2107},
  year={2010},
  
}

@article{xiao2018deepfocus,
  title={DeepFocus: learned image synthesis for computational displays},
  author={Xiao, Lei and Kaplanyan, Anton and Fix, Alexander and Chapman, Matthew and Lanman, Douglas},
  journal={ACM Transactions on Graphics},
  volume={37},
  number={6},
  pages={1--13},
  year={2018},
  publisher={ACM New York, NY, USA}
}

@article{zhang2019synthetic,
  title={Synthetic defocus and look-ahead autofocus for casual videography},
  author={Zhang, Xuaner and Matzen, Kevin and Nguyen, Vivien and Yao, Dillon and Zhang, You and Ng, Ren},
  journal={ACM Transactions on Graphics},
  volume={38},
  number={4},
  pages={1--16},
  year={2019},
  publisher={ACM New York, NY, USA}
}

@article{wadhwa2018synthetic,
  title={Synthetic depth-of-field with a single-camera mobile phone},
  author={Wadhwa, Neal and Garg, Rahul and Jacobs, David E and Feldman, Bryan E and Kanazawa, Nori and Carroll, Robert and Movshovitz-Attias, Yair and Barron, Jonathan T and Pritch, Yael and Levoy, Marc},
  journal={ACM Transactions on Graphics},
  volume={37},
  number={4},
  pages={1--13},
  year={2018},
  publisher={ACM New York, NY, USA}
}

@article{collins2017visual,
  title={Visual coherence in mixed reality: A systematic enquiry},
  author={Collins, Jonny and Regenbrecht, Holger and Langlotz, Tobias},
  journal={Presence},
  volume={26},
  number={1},
  pages={16--41},
  year={2017},
  publisher={MIT Press}
}

@inproceedings{fournier1993common,
  title={Common illumination between real and computer generated scenes},
  author={Fournier, Alain and Gunawan, Atjeng S and Romanzin, Chris},
  booktitle={Graphics Interface},
  pages={254--254},
  year={1993},
}

@article{alhakamy2020real,
  title={Real-time illumination and visual coherence for photorealistic augmented/mixed reality},
  author={Alhakamy, A’Aeshah and Tuceryan, Mihran},
  journal={ACM Computing Surveys},
  volume={53},
  number={3},
  pages={1--34},
  year={2020},
  publisher={ACM New York, NY, USA}
}

@inproceedings{chalmers2014perceptually,
  title={Perceptually Optimised Illumination for Seamless Composites.},
  author={Chalmers, Andrew and Choi, Jong Jin and Rhee, Taehyun},
  booktitle={Pacific Graphics (Short Papers)},
  year={2014}
}

@inproceedings{okumura2006augmented,
  title={Augmented reality based on estimation of defocusing and motion blurring from captured images},
  author={Okumura, Bunyo and Kanbara, Masayuki and Yokoya, Naokazu},
  booktitle={IEEE/ACM International Symposium on Mixed and Augmented Reality},
  pages={219--225},
  year={2006},
  organization={IEEE}
}

@article{klein2009simulating,
  title={Simulating low-cost cameras for augmented reality compositing},
  author={Klein, Georg and Murray, David W},
  journal={IEEE Transactions on Visualization and Computer Graphics},
  volume={16},
  number={3},
  pages={369--380},
  year={2009},
  publisher={IEEE}
}

@inproceedings{fischer2006enhanced,
  title={Enhanced visual realism by incorporating camera image effects},
  author={Fischer, Jan and Bartz, Dirk and Stra{\ss}er, Wolfgang},
  booktitle={IEEE/ACM International Symposium on Mixed and Augmented Reality},
  pages={205--208},
  year={2006},
  organization={IEEE}
}

@article{aittala2010inverse,
  title={Inverse lighting and photorealistic rendering for augmented reality},
  author={Aittala, Miika},
  journal={The Visual Computer},
  volume={26},
  pages={669--678},
  year={2010},
  publisher={Springer}
}

@inproceedings{park2009esm,
  title={ESM-Blur: Handling \& rendering blur in 3D tracking and augmentation},
  author={Park, Youngmin and Lepetit, Vincent and Woo, Woontack},
  booktitle={IEEE International Symposium on Mixed and Augmented Reality},
  pages={163--166},
  year={2009},
  organization={IEEE}
}

@inproceedings{pilet2006all,
  title={An all-in-one solution to geometric and photometric calibration},
  author={Pilet, Julien and Geiger, Andreas and Lagger, Pascal and Lepetit, Vincent and Fua, Pascal},
  booktitle={IEEE/ACM International Symposium on Mixed and Augmented Reality},
  pages={69--78},
  year={2006},
  organization={IEEE}
}

@inproceedings{kan2012physically,
  title={Physically-Based Depth of Field in Augmented Reality.},
  author={K{\'a}n, Peter and Kaufmann, Hannes},
  booktitle={Eurographics (Short Papers)},
  pages={89--92},
  year={2012}
}

@incollection{building2023,
  title={Building Statistical Models
in Python},
  author={Huy Hoang Nguyen and Paul, N Adams and Stuart, J Miller},
  booktitle={Packt Publishing},
  number={6},
  year={2023}
}

@inproceedings{ruan2022learning,
  title={Learning to deblur using light field generated and real defocus images},
  author={Ruan, Lingyan and Chen, Bin and Li, Jizhou and Lam, Miuling},
  booktitle={Proceedings of the IEEE/CVF Conference on Computer Vision and Pattern Recognition},
  pages={16304--16313},
  year={2022}
}

@article{rhee2017mr360,
  title={Mr360: Mixed reality rendering for 360 panoramic videos},
  author={Rhee, Taehyun and Petikam, Lohit and Allen, Benjamin and Chalmers, Andrew},
  journal={IEEE Transactions on Visualization and Computer Graphics},
  volume={23},
  number={4},
  pages={1379--1388},
  year={2017},
  publisher={IEEE}
}

@inproceedings{debevec1998rendering,
  title={Rendering synthetic objects into real scenes: bridging traditional and image-based graphics with global illumination and high dynamic range photography},
  author={Debevec, Paul},
  booktitle={Proceedings of the Conference on Computer Graphics and Interactive Techniques},
  pages={189--198},
  year={1998}
}

@inproceedings{sheng2024dr,
  title={Dr. bokeh: differentiable occlusion-aware bokeh rendering},
  author={Sheng, Yichen and Yu, Zixun and Ling, Lu and Cao, Zhiwen and Zhang, Xuaner and Lu, Xin and Xian, Ke and Lin, Haiting and Benes, Bedrich},
  booktitle={Proceedings of the IEEE/CVF Conference on Computer Vision and Pattern Recognition},
  pages={4515--4525},
  year={2024}
}

@article{peng2024bokehme++,
  title={BokehMe++: Harmonious Fusion of Classical and Neural Rendering for Versatile Bokeh Creation},
  author={Peng, Juewen and Cao, Zhiguo and Luo, Xianrui and Xian, Ke and Tang, Wenfeng and Zhang, Jianming and Lin, Guosheng},
  journal={IEEE Transactions on Pattern Analysis and Machine Intelligence},
  year={2024},
  publisher={IEEE}
}

@article{liu2021discrimination,
  title={Discrimination-aware network pruning for deep model compression},
  author={Liu, Jing and Zhuang, Bohan and Zhuang, Zhuangwei and Guo, Yong and Huang, Junzhou and Zhu, Jinhui and Tan, Mingkui},
  journal={IEEE Transactions on Pattern Analysis and Machine Intelligence},
  volume={44},
  number={8},
  pages={4035--4051},
  year={2021},
  publisher={IEEE}
}
\end{document}